\documentclass[lettersize,journal]{IEEEtran}
\usepackage[T1]{fontenc}
\usepackage{graphicx}
\usepackage[ruled,linesnumbered]{algorithm2e}
\usepackage{comment}
\usepackage{booktabs}
\usepackage{amssymb,amsmath,amsthm}
\usepackage{multicol}
\usepackage{multirow}
\usepackage{threeparttable}
\usepackage{cite}
\usepackage{url}
\usepackage{color}
\usepackage{diagbox}
\usepackage{arydshln} 

\begin{document}

\title{Learning to Reorient Objects with \\Stable Placements Afforded by Extrinsic Supports}

\author{
Peng Xu, Hu Cheng, Jiankun Wang$^*$,~\IEEEmembership{Senior Member,~IEEE}, and Max Q.-H. Meng$^*$,~\IEEEmembership{Fellow,~IEEE}
\thanks{$^*$Corresponding authors: Jiankun Wang, Max Q.-H. Meng.}
\thanks{Peng Xu and Hu Cheng are with the Department of Electronic Engineering, The Chinese University of Hong Kong, Hong Kong
        {\tt\small \{peterxu,hcheng\}@link.cuhk.edu.hk}}%
\thanks{Jiankun Wang is with the Department of Electronic and Electrical Engineering of the Southern University of Science and Technology in Shenzhen, China
        {\tt\small wangjk@sustech.edu.cn}}%
\thanks{Max Q.-H. Meng is with the Department of Electronic and Electrical Engineering of the Southern University of Science and Technology in Shenzhen, China, on leave from the Department of Electronic Engineering, The Chinese University of Hong Kong, Hong Kong, and also with the Shenzhen Research Institute of the Chinese University of Hong Kong in Shenzhen, China
        {\tt\small max.meng@ieee.org}}%
}

\markboth{Journal of \LaTeX\ Class Files,~Vol.~0, No.~0, February~2022}%
{Shell \MakeLowercase{\textit{et al.}}: A Sample Article Using IEEEtran.cls for IEEE Journals}

\IEEEpubid{0000--0000/00\$00.00~\copyright~2022 IEEE}

\maketitle

\begin{abstract}

Reorienting objects by using supports is a practical yet challenging manipulation task. Owing to the intricate geometry of objects and the constrained feasible motions of the robot, multiple manipulation steps are required for object reorientation. 
In this work, we propose a pipeline for predicting various object placements from point clouds. This pipeline comprises three stages: a pose generation stage, followed by a pose refinement stage, and culminating in a placement classification stage. 
We also propose an algorithm to construct manipulation graphs based on point clouds. Feasible manipulation sequences are determined for the robot to transfer object placements.
Both simulated and real-world experiments demonstrate that our approach is effective. The simulation results underscore our pipeline's capacity to generalize to novel objects in random start poses. 
{{Our predicted placements exhibit a 20\% enhancement in accuracy compared to the state-of-the-art baseline.}}
Furthermore, the robot finds feasible sequential steps in the manipulation graphs constructed by our algorithm to accomplish object reorientation manipulation.

\end{abstract}

\def\abstractname{Note to Practitioners}
\begin{abstract}

Object reorientation is a prevalent manipulation task in both domestic and industrial manufacturing scenarios. Extrinsic supporting items are often used to provide diverse object placements that allow for feasible grasp configurations for robotic manipulation.
In previous methods, utilizing mesh models of objects was necessary to ascertain stable placements and construct manipulation graphs. 
In this work, we propose a data-driven approach to predict various object placements conditioned on point clouds. 
Moreover, we use predicted point cloud placements to construct manipulation graphs, which facilitate collision-free pick-and-place steps to reorient objects. Our approach demonstrates the capacity to generalize to novel objects.
In future work, we will enhance the performance of our pipeline by optimizing the distance metric used for measuring pose discrepancies and improving the classifier model.

\end{abstract}


\section{Introduction}
%


\IEEEPARstart{I}{n} robotic manipulation, such as using tools, object reorientation is needed \cite{wan2019regrasp}. Compared to in-hand manipulation, extrinsic manipulation using supports allows simpler robot hands to reorient objects.  
\begin{figure}[tp]
      \centering
      \includegraphics[width=8.5cm]{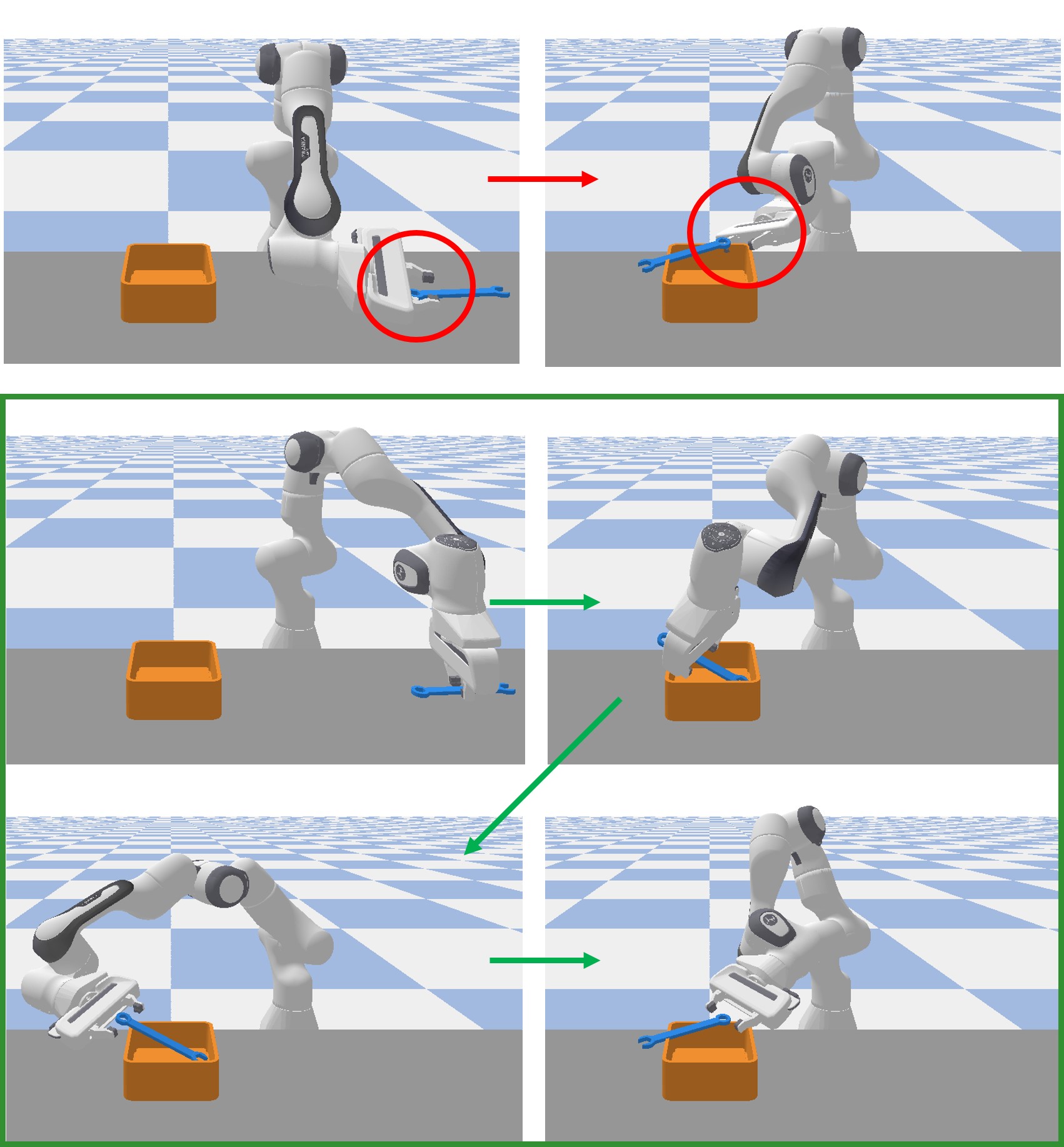}
      \caption{An example of object reorientation that requires multiple pick-and-place steps. Due to obstruction caused by the table and the tray, the robot cannot directly grasp and flip the wrench. However, utilizing the intermediate stable placement provided by the tray, the robot can first pick up the wrench and place it in the tray. Subsequently, the robot regrasps the wrench and flips it without collision. This allows the robot to use the other side of the wrench.}
      \label{f1}
   \end{figure}
Due to the kinematic constraints of the robot and the complex geometry of the object, one pick-and-place step is not enough for the robot to move the object to the target pose. Extrinsic supports can provide stable object placements that serve to facilitate feasible motions of the robot. As a result, the execution of sequential pick-and-place steps achieves object reorientation.
As illustrated in Fig. \ref{f1}, the table precludes the robot from reaching the grasp configuration in proximity to the wrench's bottom.
\IEEEpubidadjcol
Consequently, the robot arm cannot flip the wrench and place it on the support. 
To overcome this challenge, the robot initially picks up the wrench and positions it in a stable intermediate placement afforded by the tray. This placement allows access to grasp configurations that are beyond reach when the wrench is on the table.
Then the robot regrasps the wrench and achieves the flipping.

Different from in-hand manipulation methods \cite{yuan2020design,li2020learning} that rely on dexterous hands to reorient objects, extrinsic manipulation methods \cite{cao2016analyzing,ma2018regrasp} utilize stable placements provided by supporting items for feasible robot motions. The key components for reorientation manipulation are the stable placements of objects, which are supported by supporting items, and sequential pick-and-place operations for the robot to transform the placements.
Wan \textit{et al.} \cite{wan2019regrasp} proposed a method to calculate the stable placements of wooden blocks on the table based on their mesh models and build two-layer grasp graphs to connect them with calculated grasp configurations of the gripper. However, this approach is limited to objects with simple and known geometry. Ma \textit{et al.} \cite{ma2018regrasp} used the Bullet simulator to simulate and obtain stable placements of objects with complex supports. Although they built grasp graphs in parallel with the simulation process, their approach requires detailed mesh models of the objects and supports to be loaded into the simulation.
Several works \cite{haustein2019object,haustein2019placing,hou2019reorienting} have focused on pushing or pivoting objects while keeping them in the robot's hands. However, these methods can only operate or build dexterous manipulation graphs with physical attributes of objects, such as friction and masses.

Data-driven methods have shown promising results in predicting poses of unseen objects based on sensor data. 
In \cite{you2021omnihang}, two neural network models are introduced to predict hanging poses of objects. The first model takes point clouds as input and predicts the approximate pose. The second model refines the hanging pose by predicting contact points on objects. However, this approach only generates a single hanging pose for an object. In contrast, we propose a novel pipeline that incorporates neural network models to capture the distribution of diverse object placements.
Furthermore, Paxton \textit{et al.} \cite{paxton2021predicting} used point clouds of object placements to train a discriminator for scene classification. Similarly, our pipeline incorporates a classifier model to discriminate different placements.

Our contributions to the field of extrinsic manipulation for object reorientation are outlined as follows: 
(1) Our approach demonstrates the capability to predict various placements with higher accuracy, surpassing the existing state-of-the-art method. {{Experiment results show that the overall accuracy of our approach is 20\% higher than the previous work}}.
(2) We propose an algorithm to construct manipulation graphs based on point clouds. Our approach enables the robot to reorient previously unseen objects with sequential pick-and-place steps. 



\section{Related Work}

Reorienting objects can be achieved in two main ways. The first is to use dexterous hands and a control policy to adjust the object's pose without extrinsic supports. Previous works on in-hand manipulation \cite{li2020learning, yuan2020design, sundaralingam2018geometric} have incorporated dexterous end-effectors and a manipulation policy obtained through imitation learning \cite{yuan2020design}, reinforcement learning \cite{li2020learning}, or optimization methods \cite{sundaralingam2018geometric} to adjust the object's position.
The second way is to use placement poses provided by extrinsic supporting items, which can be as simple as a desktop \cite{wan2019regrasp} or as complex as a container \cite{ma2018regrasp} or another gripper \cite{haustein2019object, haustein2019placing}. In this case, the robot transforms the object's pose. Several works \cite{wan2019preparatory, cruciani2019dual, haustein2019placing, haustein2019object, wan2016achieving} have proposed methods for reorienting objects with two robot arms, where the poses of objects are adjusted in mid-air with the help of the supporting gripper. For example, \cite{wan2019preparatory} hands over the object between two grippers without changing the object's pose relative to the grippers, while \cite{haustein2019placing, haustein2019object, cruciani2019dual} push the object grasped in one end-effector using another end-effector. These methods involve sequential operations for adjusting the object's pose. Therefore, we also discuss relevant works on manipulation graph construction.

\begin{figure*}[t]
\centering
\includegraphics[width=\linewidth]{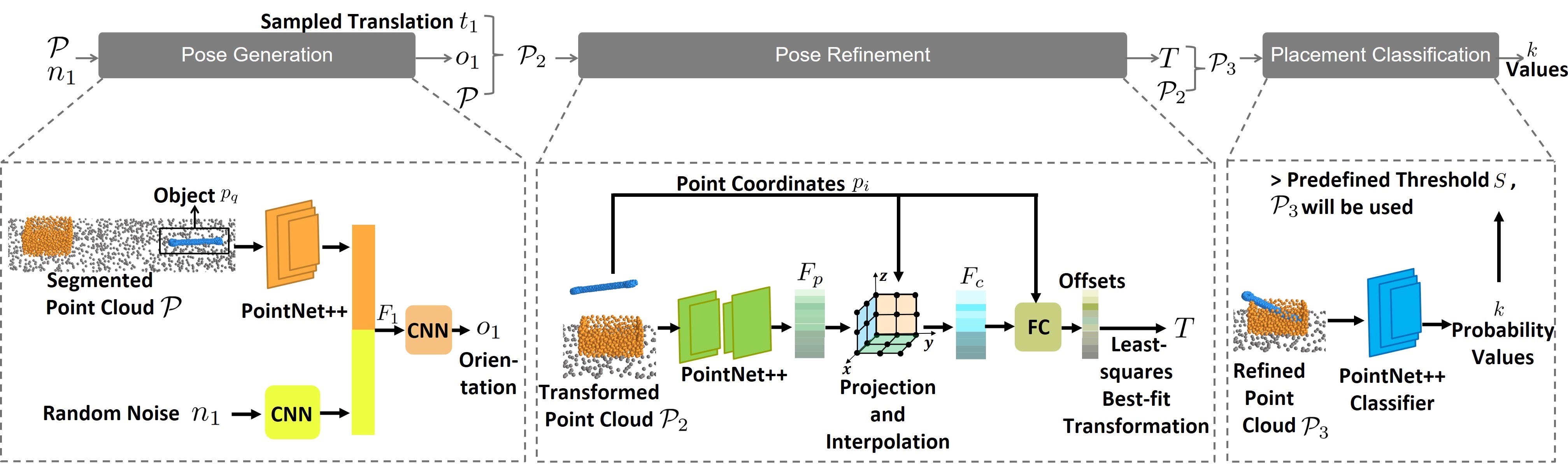}
\caption{{{An overview of our pipeline. We leverage a three-stage pipeline to predict and classify object placements.}} }
\label{f2}
\end{figure*}

Wan \textit{et al.} \cite{wan2015reorientating,wan2019regrasp} propose a method for adjusting the poses of objects on a flat table surface using a single robot arm, given mesh models of objects and the gripper. They calculate the convex hulls of the query objects and compute the objects' stable placements on the table.
Hou \textit{et al.} \cite{hou2019reorienting,Hou-2018-105030} adopt pivoting, a manipulation primitive that relies on the gripper model and contact constraints, for object reorientation. The objects contact the gripper and the working platform in the pivoting process, and mesh models are required for mechanical analysis.
Some works consider complex supports for obtaining stable placements of objects. Cao \textit{et al.} \cite{cao2016analyzing} use a vertical pin to support the query object and compute stable placements that differ from poses supported by a horizontal surface. However, they focus on relatively simple supporting items. Obtaining objects' stable placements on complex supports is challenging. Ma \textit{et al.} \cite{ma2018regrasp} use the Bullet dynamic simulator to obtain the object's stable placements on a complex support through free drop trials in simulation. However, this approach is time-consuming and requires detailed physical information for accurate simulation results, such as the masses of models and friction.

Several works have explored data-driven methods for generalizing to novel objects and obtaining stable placements on complex supports based on visual perception. Jiang \textit{et al.} \cite{jiang2012learning,jiang2012learning2} manually design features to represent point clouds of objects and use Support Vector Machines to obtain object placements on several supports. Berscheid \textit{et al.} \cite{berscheid2020self} propose a constructive learning method that considers both stable and unstable poses to obtain placement poses for various objects. Paxton \textit{et al.} \cite{paxton2021predicting} solve the rearrangement task to place objects based on semantic instructions and use a discriminator trained on stable and unstable poses to classify different placements. They also use an optimization approach to determine the placement pose of the query object that satisfies a specific instruction. Similarly, we train a classifier to discriminate different placements. We select stable placements with high stability scores assigned by our classifier.
Other works, such as Cheng \textit{et al.} \cite{cheng2021learning} and You \textit{et al.} \cite{you2021omnihang}, use neural network models that predict contact points on the objects to generate object placements. These approaches consider the contact between the object and the support while ignoring the table. However, more than a single placement is needed for object reorientation, and the robot's motions need to be considered.

In addition to inferring stable placements of objects, the key to implementing object reorientation is the motion of the robot to transform these placements. 
Simeonov \textit{et al.} \cite{simeonov2020long} propose a framework for long-term planning of object manipulation, which jointly models reorientation subgoals and contact configurations to achieve subgoals.
Many works achieve object reorientation through sequential operations, constructing manipulation graphs using intermediate object placements and the calculated grasp configurations of the robot to reorient the objects in a collision-free manner.
For instance, after calculating stable placements of objects on the table, Wan \textit{et al.} \cite{wan2019regrasp} built a two-layer regrasping graph, where the first layer shows connectivity between two stable placements and the second layer shows shared grasp configurations. Meanwhile, Ma \textit{et al.} \cite{ma2018regrasp} built the regrasping graph in parallel with the simulation process of obtaining the object's stable placements. However, the simulation process takes longer to obtain more stable placements that are added to the manipulation graph.
Other works, such as Haustein \textit{et al.} \cite{haustein2019placing,haustein2019object}, move the object to a new pose by pushing the object in-hand as per the dexterous manipulation graph with optimized grasp configurations. In \cite{Hou-2018-105030,hou2019reorienting}, motion primitives such as rolling, pivoting, and pick-and-place are used to construct manipulation graphs. Nonetheless, these approaches require detailed mesh models of objects.
In our work, we compute grasp configurations based on point clouds. We take the extrinsic support as the supporting item and establish the world coordinate frame at its bottom to simplify the computation.

\section{Approach}


\subsection{Approach Overview}

We propose a solution to reorient objects using placements afforded by extrinsic supports. Our proposed solution comprises two main components: a pipeline that predicts placements based on point clouds and an algorithm that constructs manipulation graphs with point clouds. 

Specifically, we employ neural networks to learn the object’s stable poses from point clouds that are fused from multiple viewpoints. 
Multi-view fusion is a mature technique  \cite{breyer2020volumetric} that results in more points than those obtained from a single shot, eliminating the hassle of reconstructing point clouds. 
Our approach assumes that the two objects are separate, with initial relative positions that are arbitrary. 
We employ the Point Cloud Library \cite{Rusu_ICRA2011_PCL} to perform segmentation of the fused point cloud, partitioning it into a plane and two distinct objects. Each point within the segmented point cloud is assigned a label in an additional dimension to differentiate between objects. Subsequently, the point cloud coordinates, in conjunction with their respective labels, are passed into our pipeline.
We denote the initial input point cloud as $\mathcal{P} \in \mathbb{R}^{2048 \times 4}$. We denote the object to be manipulated as $p_{q}$.
The neural network models that we used are explained in detail in Sec. \ref{sec3.1}, while the dataset used for training the models is presented in Sec. \ref{sec3.3}. 


Our algorithm computes shared grasp configurations to connect placements generated by our pipeline. The detailed construction process of the manipulation graphs is discussed in Sec. \ref{sec3.2}.

\subsection{The Pipeline for Placement Prediction}
\label{sec3.1}

Fig. \ref{f2} provides an overview of our pipeline. 
Our pipeline entails three stages: pose generation, pose refinement, and placement classification. These stages are interconnected, where the output of the first stage is the input of the second stage, and so on.
Firstly, the input point cloud $\mathcal{P}$ is used to generate orientations of placements. The generated orientations are utilized for transformation. Next, the transformed point clouds $\mathcal{P}_{2}$ are refined to ensure they are accurate for physical contact between objects. Finally, the refined placements $\mathcal{P}_{3}$ are classified to distinguish stable placements, which are then used to construct manipulation graphs.




\subsubsection{Pose Generation}
\label{secPG}
The primary objective of the first stage is to predict diverse orientations of objects based on the input point cloud $\mathcal{P}$. 
{{To capture distributions over object placements conditioned on visual perception, we use the generator network from the variational autoencoder \cite{veres2017modeling}. The generator takes an estimate of the posterior and learns the data distribution. Typically, the posterior is a multivariate Gaussian.}} 
Our model also takes as input random noise sampled from a multivariate normal distribution $\mathcal N(\vec\mu,\sigma^2 \textbf{I})$ that is parameterized by a zero mean vector $\vec\mu$ and a covariance matrix $\sigma^2 \textbf{I}$. 
We employ a PointNet++ \cite{qi2017pointnet++} with set abstraction layers to encode the input point cloud $\mathcal{P}$ into a feature vector and a convolutional neural network (CNN) to encode a random sample $n_{1}$ into another feature vector. These resultant feature vectors are concatenated to form a single feature vector $F_{1}$, which is then passed into a CNN that maps $F_{1}$ into an orientation $o_{1}$. 

During training, ${M}$ random samples are introduced so that the model outputs a set of orientations. 
Specifically, we input the point cloud $\mathcal{P} \in \mathbb{R}^{2048 \times 4}$ and random samples $\{ n \} \in \mathbb{R}^{3 \times M}$, and the model generates orientations $\{ o \} \in \mathbb{R}^{3 \times M}$. 
The set of orientations  $\{ o \}$ is regarded as a point set. The set of ground-truth orientations $\{  o_{gt} \}$ of diverse placements that are collected in simulation is also a point set. We use Chamfer Distance \cite{fan2017point} to describe the difference between the two sets. Thus, the loss function is defined as follows:
\begin{equation}
\mathcal{L}_{gen} \\=\\ \sum_{x \in \{ o \}}\\ \min_{y \in \{  o_{gt} \}} \\{{||x-y||}_{2}^{2}} \\+\\ \sum_{y \in \{  o_{gt} \}}\\ \min_{x \in \{ o \}} \\{{||x-y||}_{2}^{2}}.
\label{eq.1}
\end{equation}

During pose inference, our model generates the orientation that is used to rotate the object $p_{q}$. The translation $t_{1}$ is sampled based on the bounding box of the support to ensure that the object is positioned in the vicinity of the support. Specifically, the $x$ and $y$ coordinates are sampled from a normal distribution with a mean of zero and a standard deviation proportional to the bounding box size. The $z$ coordinate is higher than the height of the support to mitigate penetration. This way, we can obtain point clouds like $\mathcal{P}_{2}$.

\subsubsection{Pose Refinement}
\label{secPR}
To refine the coarse placements of objects, the prior method \cite{cheng2021learning} has relied on predicting the points on the object and contact with the support to adjust the placements. However, this method is prone to producing penetration cases due to imprecise contact point predictions. 
{{In \cite{shen2022acid}, implicit representations have demonstrated potential in modeling dynamics in high fidelity. Inspired by \cite{shen2022acid}, which learns structured implicit representations of the dynamics field from visual perception inputs, we propose a coordinate-based model to learn implicit representations of forward dynamics under gravity to tackle the problem of pose refinement. Our model consists of two main components: an encoder for mapping point clouds into canonical feature planes and a decoder for mapping coordinates and their corresponding queried features into forward dynamics offsets.}}

Firstly, we use a PointNet++ network with set abstraction layers and feature propagation layers to map the transformed point cloud $\mathcal{P}_{2}$ into per-point feature vectors $F_{p}$, considering local information of the inputs. 
{{Following the projection operations outlined in \cite{shen2022acid}, we orthographically project these feature vectors onto three canonical feature planes, utilizing normalized input point coordinates. These feature planes are then passed through U-Nets \cite{ronneberger2015u} that consist of a sequence of down- and upsampling convolutional layers. The obtained feature planes are implicit representations of the dynamics field. 
According to \cite{peng2020convolutional}, the model is endowed with translation equivariance by employing the projection operations followed by the use of U-Nets. Furthermore, the feature planes are inpainted by the convolutions. These encoding operations address the major limitation posed by the fully connected network architecture of previous implicit methods, which fails to integrate local information from inputs and incorporate translation equivariance into the model, ultimately hindering the structured reasoning of dynamics.}}

We query the feature at a point $p_i \in \mathcal{P}_{2}$ coordinates by aggregating  the features computed through bilinear interpolation from each feature plane. Finally, multiple fully-connected (FC) layers are used to map the point coordinates and their corresponding queried features $F_{c}$ into forward dynamics offsets under gravity. 
We employ the mean squared error (MSE) to calculate the difference between the offsets of input points under gravity and the corresponding offsets predicted by our model. The loss function is expressed as follows:
\begin{equation}
\mathcal{L}_{ref} = \frac{\sum_{i = 1}^{N} MSEloss( p_{i}^{gt} - p_i, t_i )}{N}, 
\label{eq.5}
\end{equation}
where $p_{i}^{gt}$ is the coordinate of an input point at the stable placement, $p_i$ denotes the coordinate of this input point ($i \leq N$), and $t_i$ denotes the predicted offset for this point. $N$ is the number of points considered for computation, including both object points and support points.
A new set of object points is obtained by adding predicted offsets to the input object point cloud.
To preserve the shape of the object after pose refinement, we calculate the least-squares best-fit transformation $T$ that maps the object’s points to the new point set. An exemplary result using the transformation $T$ is shown as $\mathcal{P}_{3}$.

\subsubsection{Placement Classification}

In the third stage, we employ a classifier neural network model to categorize the point clouds with refined poses. The placements of the object afforded by the support belong to different types, each characterized by its spatial configuration. Similarly, the unstable relative positions between the object and the support contain various types subject to spatial configurations. 
Our classifier aims to not only distinguish the stability of the input point clouds $\mathcal{P}_{3}$ but also categorize their types. We utilize a PointNet++ model with set abstraction layers to encode the input point cloud into a feature map. Subsequently, a multi-layer perceptron (MLP) is used to map the feature map into $k$ probability values for $k$ categories. 
Here, the value of $k$ is 6. The detailed categorization of stable and unstable placements is presented in Sec. \ref{sec3.3}.
We use the Cross-Entropy loss to supervise the classifier. To enhance the accuracy of placements used in the manipulation process, we adopt the placements with a probability value of a stable type above the predefined threshold $S$ to construct manipulation graphs.


\begin{figure*}[ht]
      \centering
      \includegraphics[width=\linewidth]{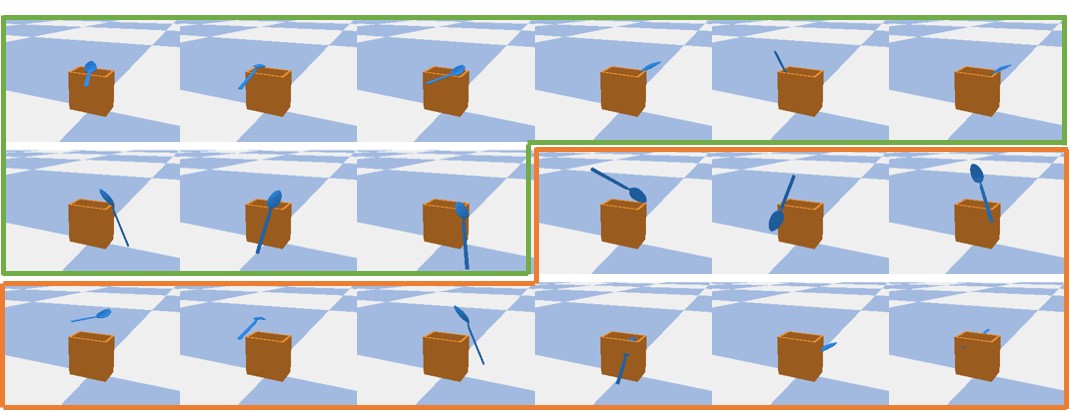}
      \caption{Examples of stable and unstable placements of a pair from the training set. The stable placements are enclosed by the green lines. The unstable placements are enclosed by the red lines.}
      \label{f4}
\end{figure*}

\subsection{Dataset}
\label{sec3.3}

We generated a large-scale dataset of object placements afforded by supports using Pybullet \cite{coumans2020}. Mesh models of objects and supports are collected from 3D Warehouse\footnote{https://3dwarehouse.sketchup.com/}, examined for convexity, resized to regular sizes, and loaded into Pybullet. The dataset consists of 35 supports and 136 commonly seen objects in domestic and industrial scenes. 
We combined them into 786 matched pairs, of which 623 pairs are used for training, 149 pairs for validation, and 14 pairs for testing. The models in the training, validation, and test sets were mutually exclusive. {{More samples of the supports and objects are available at the link\footnote{https://drive.google.com/drive/folders/1LTkSSSwRLaVtQaKNsPCaTL9x\\DIDDGfQV?usp=sharing}.}}


\begin{algorithm}

\caption{Manipulation graph construction}
\label{graph}
\KwIn{the input point cloud $\mathcal{P} \in \mathbb{R}^{2048 \times 4}$, the model of the default gripper $\mathcal{M}$, and the given goal placement $\mathcal{P}^{F}$.}
\KwOut{the manipulation graph $\mathbb{G}$}
$\mathbb{G}$.\emph{init}$(\mathcal{P}$, $\mathcal{P}^{F})$ \\
${\mathbb{P}}\leftarrow$ $\emptyset$, ${\mathbb{P}}$.\emph{append}$(\mathcal{P}$, $\mathcal{P}^{F})$\\

$\mathbb{P}_{gen}\leftarrow$ \emph{OurPipeline}$(\mathcal{P})$\\
$\mathbb{P}_{gen} \leftarrow \emph{sorted}(\mathbb{P}_{gen})$ \\

\For{$\mathcal{P}^{i}$ in $\mathbb{P}_{gen}$} 
{
$\mathcal{E}\leftarrow$ $\emptyset$\\
\If{not Connect$(\mathcal{P}$, $\mathcal{P}^{F})$}{

${\mathcal{E}}\leftarrow$ \emph{CalculateGrasp}$(\mathcal{P}^{i}, \mathbb{P}, \mathcal{M})$\\
%
%
%
$\mathbb{P}$.\emph{append}$(\mathcal{P}^{i})$\\
$\mathbb{G}$.\emph{addNodes}$(\mathcal{P}^{i})$\\
$\mathbb{G}$.\emph{addEdges}$(\mathcal{E})$\\

}

}

Return($\mathbb{G}$)\\

\end{algorithm}


For each matched pair, we let the object fall under gravity in numerous random poses above the support and simulated each process for 10 seconds. We recorded the stable placements of the object, which were afforded by the support, and examined for velocity and contact. 
Additionally, we designed classification criteria to categorize these simulated results, where objects placed on top of supports form one category, objects placed inside supports form another category, and objects supported jointly by both the support and a tabletop form a third category. 
The transformations from initial poses to stable placements were calculated in the world coordinate frame. Diverse stable placements were used to train the model in the pose generation stage. 
We also distinguished three types of unstable placements: objects without contact with the supports, objects in contact with the supports but in unstable positions, and objects penetrating through the supports. Examples of a pair's stable and unstable placements from the training set are shown in Fig. \ref{f4}. 
We recorded the unstable placements contact with supports in simulation associated with their final stable placements for use in the pose refinement stage. The other two kinds of unstable placements are obtained by changing the translation coordinates of the recorded stable placements. Specifically, the penetrating unstable placements are obtained by the downward translation of the stable placements, whereas the non-contact unstable placements result from the upward translation of the stable placements. 
We recorded a total of 12,459 stable and 37,298 unstable placements to train our classifier to classify different stable and unstable placements.

\subsection{Training Details}
The neural network models in our pipeline are implemented using PyTorch 1.6.0. We use the Adam optimizer for all the models. Our pipeline is trained on an NVIDIA 2080Ti GPU. The training curves of all the models on validation sets are converged. 
The number of input noises is 128 for ${M}$ in the pose generation stage.


\subsection{The Algorithm for Manipulation Graph Construction}
\label{sec3.2}

\begin{figure*}[ht]
\centering
\includegraphics[width=\linewidth]{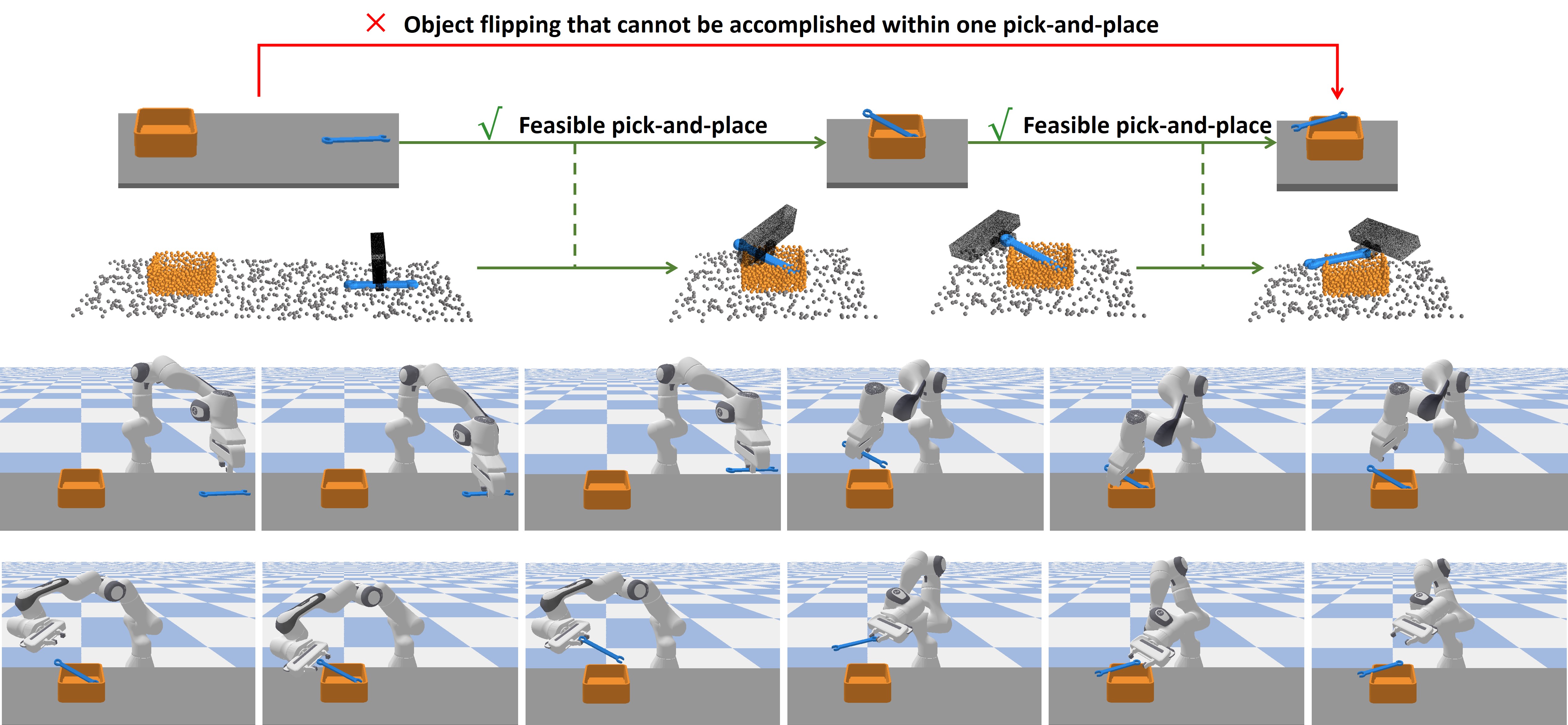}
\caption{An example of the manipulation graph. The green lines connecting the placements in the graph represent different shared grasp configurations. On each line, the relative position between the gripper and the object remains constant. The snapshots illustrate the execution of flipping an object using the defined motion primitives of pick-up $g^{up}_{xy}$ and place-down $g^{dn}_{xy}$.}
\label{f3}
\end{figure*}


In addition to predicting stable placements of objects, it is equally critical to realize collision-free placement transformations when a robot uses extrinsic supports to reorient objects. We propose an algorithm, outlined in Alg. \ref{graph}, to construct manipulation graphs based on point clouds, enabling feasible pick-and-place operations for placement transformation. Shared grasp configurations that connect placements in the manipulation graph are calculated through our algorithm. Furthermore, we incorporate motion primitives to facilitate robot motion \cite{wan2019regrasp}. We provide implementation details as follows.

Our algorithm takes the initial point cloud $\mathcal{P}$  and the given goal placement $\mathcal{P}^{F}$ as input. The object in its initial state ${p}_{q}$ cannot be moved to the goal state by a single pick-and-place of a robot, which motivates the need for our algorithm. The given goal placement serves a useful purpose, for example, as a flipped placement of the initial point cloud that enables a robot to use the opposite side of the object.  The model of the robot arm and the defaulted parallel-finger gripper $\mathcal{M}$ are assumed as given. 

In the beginning, the manipulation graph $\mathbb{G}$ is initialized with $\mathcal{P}$ and $\mathcal{P}^{F}$ (line 1). The collection of placements in the manipulation graph is denoted as the set ${\mathbb{P}}$, which is initially appended with $\mathcal{P}$ and $\mathcal{P}^{F}$ (line 2). Our pipeline of predicting placements, introduced in Sec. \ref{sec3.1}, is represented by the $\emph{OurPipeline}$. Based on the input point cloud $\mathcal{P}$ , we leverage our pipeline to generate diverse placements, each associated with a corresponding probability value (line 3). We then apply a descending sorting operation to the generated placements based on the assigned scores, thus producing an ordered set of placements $\mathbb{P}_{gen}$ (line 4).
Concretely, we obtain a set  $\mathbb{P}_{gen} = \{{\mathcal{P}}^{1}, {\mathcal{P}}^{2}, ..., {\mathcal{P}}^{Q}\}$ of $Q$ point clouds, where each entry $\mathcal{P}^{i}$ is a combination of the initial supporting environment point cloud and the transformed object point cloud with a refined transformation ${\rm \textbf T}^{i}_{q}=[{\rm \textbf R}^{i}_{q}|{\rm \textbf t}^{i}_{q}]$. 
Next, we use these placements sorted in descending order of probability values to compute feasible grasp configurations for the robot. 


The $\emph{CalculateGrasp}$ computes grasp configurations, which consist of shared grasping points on placements for the gripper, the gripper’s approaching directions to grasp the object, and the feasible kinematics of the robot. Importantly, the indices of object points remain invariant after applying a transformation ${\rm \textbf T}^{i}_{q}$. Therefore, the indices of the transformed points ${\rm \textbf T}^{i}_{q}{p}_{q}$ remain the same as $p_{q}$ in the initial point cloud. To determine shared grasping points on every two placements, we consider pairs of grasping points predicted by \cite{shao2020unigrasp} and examine grasping force closure with the friction coefficient $f$ at each pair of predicted grasping points $({q}_{x},{q}_{y})$.
The gripper's approaching directions ${\rm \textbf d}_{xy}$ with respect to the world coordinate frame are uniformly sampled at the grasping points $({q}_{x},{q}_{y})$ on a placement. In addition, the approaching directions ${\rm \textbf d}^{i}_{xy}$ and ${\rm \textbf d}^{j}_{xy}$ to grasp placements satisfy 
\begin{equation}
{\rm \textbf d}^{j}_{xy} = {\rm \textbf R}^{j}_{i} \cdot {\rm \textbf d}^{i}_{xy}
\label{eq.6}
\end{equation}
where ${\rm \textbf R}^{j}_{i}$ is the rotation matrix between ${\rm \textbf T}^{i}_{q}{p}_{q}$ and ${\rm \textbf T}^{j}_{q} {p}_{q}$ with respect to the world coordinate frame. This implies that the two placements share the same grasp configuration, and the relative positions of the two placements to the gripper remain the same.
We further examine the reachability of the grasp configuration $g_{xy}:(({q}_{x},{q}_{y}),{\rm \textbf d}_{xy})$ on two placements by checking for collisions between the gripper $\mathcal{M}$ and the supporting environment. To ensure feasible kinematics of the robotic arm, we restrict the gripper's approaching directions such that ${\rm \textbf d}_{xy} \cdot {\rm \textbf n} \leq \delta$, where ${\rm \textbf n}$ is the normal vector of the table surface.
Finally, we add force-closure and feasible grasp configurations to the manipulation graph after examining the kinematics of the entire robotic arm.
To visualize the grasp configurations, we render them in the segmented point clouds, with the gripper located at the calculated grasp. 

%


\begin{table*}[ht]

\caption{Experiment Results}
\label{table1}
\centering
\small

\resizebox{\linewidth}{!}{
\begin{tabular}{*{15}{c}}
\toprule
\multirow{2}*{\diagbox{Methods}{Pairs}} 
& \multicolumn{2}{c}{basket5+keyboard7} 
& \multicolumn{2}{c}{tray102+pliers6}
& \multicolumn{2}{c}{tray102+wrench4}  
& \multicolumn{2}{c}{plate1+phone1} 
& \multicolumn{2}{c}{plate1+phone5} 
& \multicolumn{2}{c}{plate1+wrench4}
& \multicolumn{2}{c}{plate1+pliers6} \\
\cmidrule(lr){2-3}\cmidrule(lr){4-5}\cmidrule(lr){6-7}\cmidrule(lr){8-9}\cmidrule(lr){10-11}\cmidrule(lr){12-13}\cmidrule(lr){14-15}
 &num&rate(\%)&num&rate(\%)&num&rate(\%)&num&rate(\%)&num&rate(\%)&num&rate(\%)&num&rate(\%) \\
\midrule

ours w/o refine 
&2 &14.3 
&0 &0
&0  &0 %
&24 &20.9  
&16  &15.5 %
&$\textbf{10}$ &$\textbf{11.2}$ %
&\textbf{54} &50.9 \\

\hdashline[0.5pt/5pt]
\textbf{ours}  
&$\textbf{13}$   &$\textbf{14.4}$ 
&$\textbf{12}$  &$\textbf{24.0}$
&$\textbf{17}$  &$\textbf{22.4}$
&21   &36.2 %
&15   &$\textbf{24.6}$   %
&6    &8.9 %
&44   &\textbf{53.0}  \\

\hdashline[0.5pt/5pt]
ours w/ 6d 
&6   &9.1
&6  &18.8
&7  &16.3
&1   &3.8   %
&1    &10.0 %
&2   &9.5 %
&4   &14.3  \\

L2P\cite{cheng2021learning} 
&0  &0  
&0  &0 
&10   &2.6
&$\textbf{289}$   &$\textbf{58.3}$
&$\textbf{31 }$   &8.0
&0    &0
&2    &0.5 \\

\end{tabular}

}

\resizebox{\linewidth}{!}{
\begin{tabular}{*{15}{c}}
\toprule
\multirow{2}*{\diagbox{Methods}{Pairs}} 
& \multicolumn{2}{c}{plate5+phone1} 
& \multicolumn{2}{c}{plate5+phone5} 
& \multicolumn{2}{c}{plate5+wrench4} 
& \multicolumn{2}{c}{plate5+pliers6} 
& \multicolumn{2}{c}{cup1+phone1} 
& \multicolumn{2}{c}{cup1+phone5}
& \multicolumn{2}{c}{cup1+wrench4} \\

\cmidrule(lr){2-3}\cmidrule(lr){4-5}\cmidrule(lr){6-7}\cmidrule(lr){8-9}\cmidrule(lr){10-11}\cmidrule(lr){12-13}\cmidrule(lr){14-15}
 &num&rate(\%)&num&rate(\%)&num&rate(\%)&num&rate(\%)&num&rate(\%)&num&rate(\%)&num&rate(\%) \\
\midrule

ours w/o refine 
&25  &\textbf{55.6}
&75 &55.1
&17 &56.7
&\textbf{37}  &\textbf{52.1}
&0 &0
&0 &0
&0  &0 \\

\hdashline[0.5pt/5pt]
\textbf{ours}  
&23  &40.4
&64  &\textbf{64.0}
&13  &\textbf{48.1}
&29    &46.8
&\textbf{34}  &\textbf{47.2} 
&\textbf{38}  &\textbf{42.7} 
&\textbf{12}  &\textbf{37.5}  \\

\hdashline[0.5pt/5pt]
ours w/ 6d 
&10   &\textbf{55.6}
&5  &38.5
&5 &38.5
&10  &40.0
&4   &15.4
&0    &0
&0   &0  \\

L2P\cite{cheng2021learning} 
&\textbf{212}  &44.4
&\textbf{82}  &22.5
&\textbf{53}  &17.1
&26  &6.8
&0 &0
&0 &0
&0 &0 \\

\bottomrule


\end{tabular}

}

\raggedright
\footnotesize{
   \vspace{2mm}
   
   We compare our method with its variants and the baseline L2P \cite{cheng2021learning}. The accuracy rate and the number of generated stable placements are reported in this table. The best results are in bold. 
   
   }

\end{table*}

To expand the manipulation graph $\mathbb{G}$ and achieve the goal of connecting $\mathcal{P}$ and $\mathcal{P}^{F}$ using intermediate placements, we compute the shared grasp configurations $\mathcal{E}$ between $\mathcal{P}^{i}$ that ranked by its probability value, and each placement in the set ${\mathbb{P}}$ (line 8). 
Then, the placement $\mathcal{P}^{i}$ is appended into the set ${\mathbb{P}}$ for the calculation of grasp configurations shared with subsequent placements to enhance the graph’s capability to provide manipulation operations (line 9). We add $\mathcal{P}^{i}$ and the corresponding grasp configurations $\mathcal{E}$ into the manipulation graph (lines 10-11). If $\mathcal{P}$ and $\mathcal{P}^{F}$ are not connected via $\mathcal{P}^{i}$, we perform computations of grasp configurations considering new placements until the connection is achieved in the manipulation graph (lines 5-13). Ultimately, we return the graph $\mathbb{G}$, providing the robot with comprehensive grasp configurations for reorienting objects. 

The grasp configurations are established in manipulation graph construction. To facilitate the implementation of pick-and-place, we also design motion primitives similar to those in previous work  \cite{wan2019regrasp}. Specifically, we assume that the robot moves the gripper to a position above the object before grasping it, and similarly, before placing the object, the robot moves it to a position above the target placement. These operation expressions are defined as pick-up $g^{up}_{xy}$ and place-down $g^{dn}_{xy}$:
\begin{equation}
g^{up}_{xy}=((g_{xy}, \varepsilon, h) \rightarrow (g_{xy}, \varepsilon) \rightarrow g_{xy}),\ 
\label{eq.7}
\end{equation}
\begin{equation}
g^{dn}_{xy}=((g_{xy}, h) \rightarrow g_{xy} \rightarrow (g_{xy}, \varepsilon, h)),
\label{eq.8}
\end{equation}
where $g_{xy}$ is the grasp configuration calculated by $\emph{CalculateGrasp}$, $ \varepsilon$ is the distance between two fingers of the default gripper when it is open, and $h$ is the height above the targets. The $g^{up}_{xy}$ describes the process of the robot moving above the object with long finger distance $(g_{xy}, \varepsilon, h)$, approaching the object $(g_{xy}, \varepsilon)$, and grasping the object $g_{xy}$. The $g^{dn}_{xy}$ describes the process of the robot moving the object above the target placement $(g_{xy}, h)$, placing the object down $g_{xy}$, and releasing the object $(g_{xy}, \varepsilon, h)$.
As depicted in Fig. \ref{f3}, the gripper approaches the targets from above before grasping the object or placing it down. These motion primitives allow to relax the motion planning process. Additionally, we ensure these motion primitives are collision-free and feasible with inverse kinematics.


\section{Experiments}
\label{sec5}


We evaluate our pipeline and algorithm on unseen pairs of objects and supports. We compare our pipeline with the baseline method \cite{cheng2021learning} in terms of both diversity and accuracy of predicted placements. Furthermore, we conduct ablation experiments to demonstrate the contribution of the main stages in our pipeline. We also discuss the limitations of our pipeline. To validate our algorithm for constructing manipulation graphs, we showcase a robot flipping unseen objects in the simulation. Finally, we provide a real robot experiment to demonstrate the effectiveness of our approach. 


\begin{figure*}[t]
      \centering

\includegraphics[width=\linewidth,trim=0 33 0 38,clip]{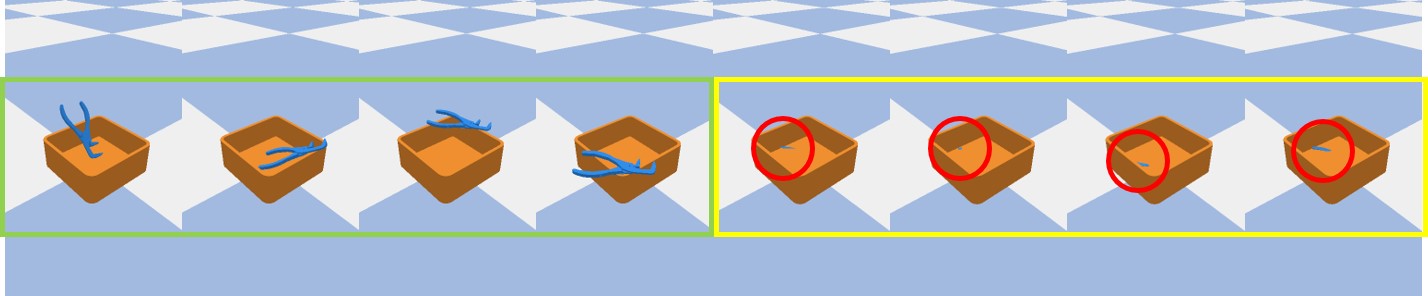}

\includegraphics[width=\linewidth,trim=0 31 0 45,clip]{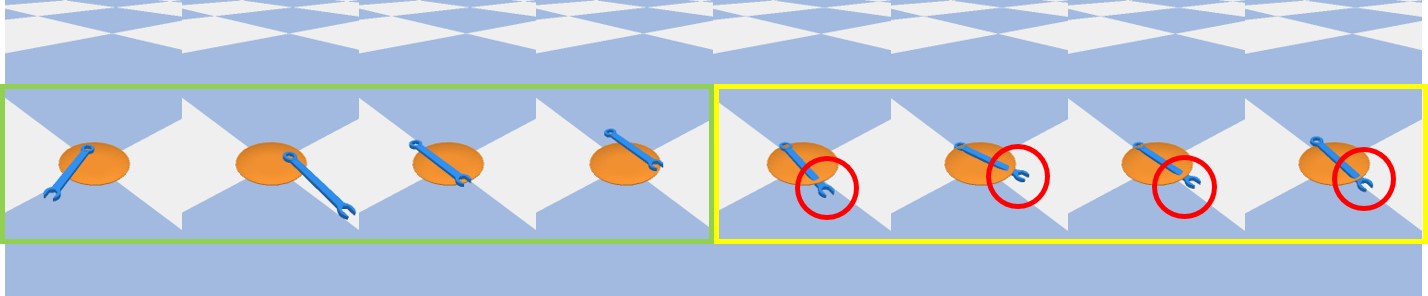}

\includegraphics[width=\linewidth,trim=0 31 0 50,clip]{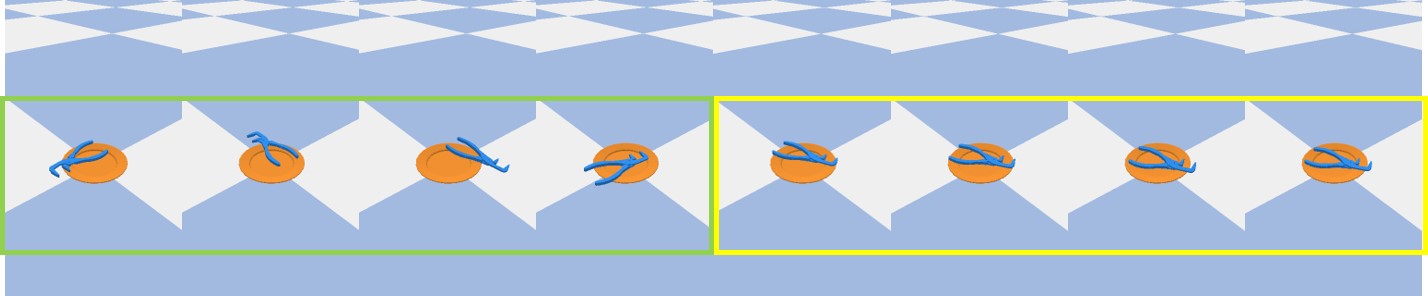}

\includegraphics[width=\linewidth,trim=0 26 0 45,clip]{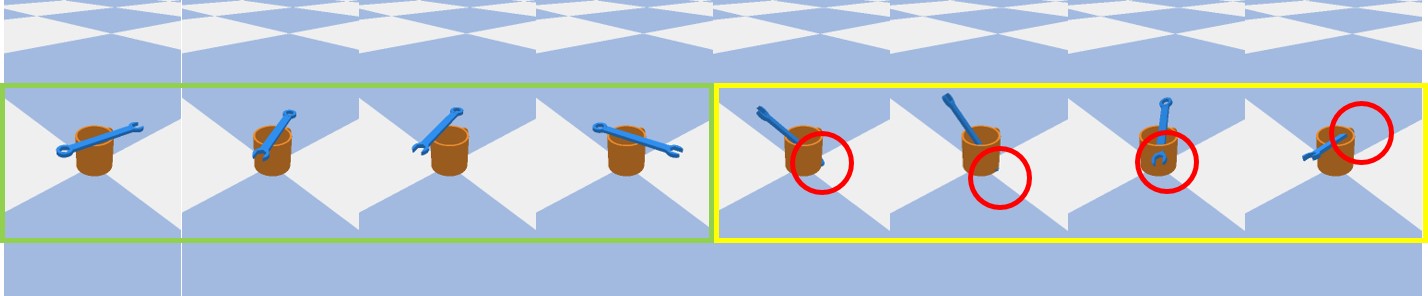}

      \caption{Comparison between placements generated by our approach and the baseline. The results of tray102+pliers6, plate1+wrench4, plate5+pliers6, and cup1+wrench4 are displayed in the simulation environment. The placements predicted by our approach are in the green box. The placements generated by the baseline are in the yellow box. Penetration cases are marked in red circles.
      }
      \label{f9}
      
\end{figure*}

\begin{figure}[ht]
      \centering
\includegraphics[width=\linewidth,trim=0 20 0 38,clip]{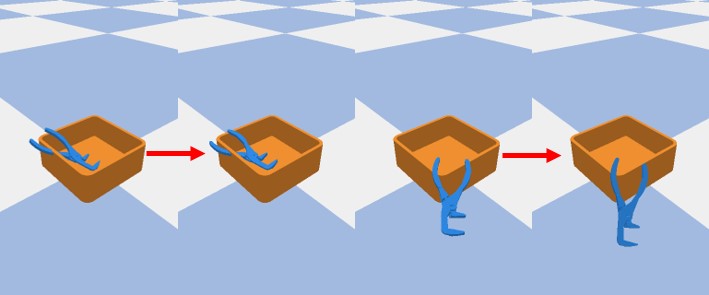}

\includegraphics[width=\linewidth,trim=0 31 0 45,clip]{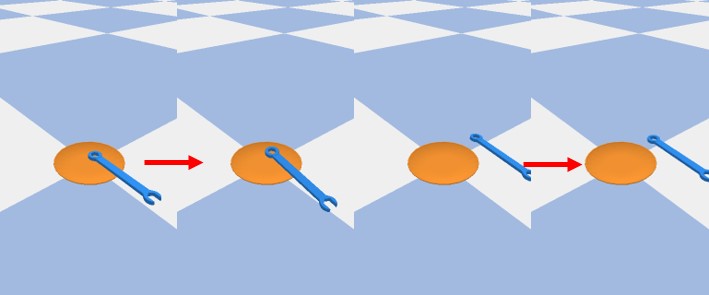}

\includegraphics[width=\linewidth,trim=0 20 0 45,clip]{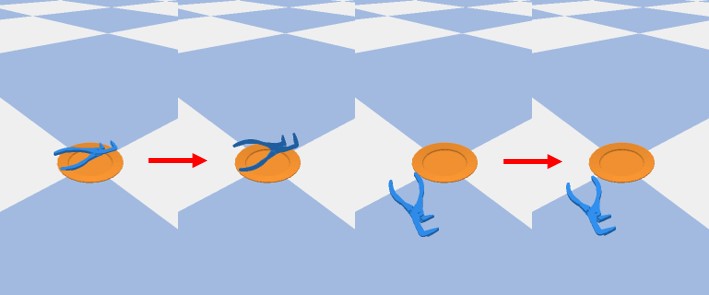}

\includegraphics[width=\linewidth,trim=0 35 0 40,clip]{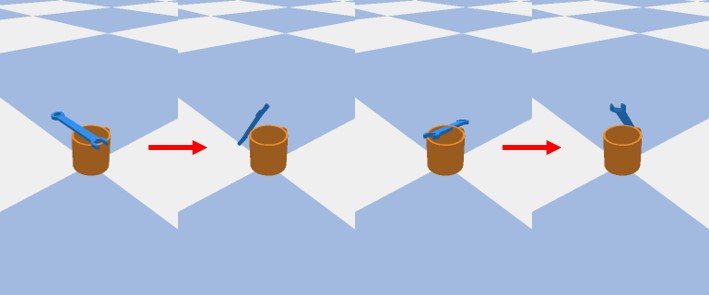}
      
      \caption{Failure examples of placements generated by our approach. The generated placements and their simulated positions under gravity are shown.}
      \label{failure}
   \end{figure}










\begin{figure*}[!htb]
      \centering
      
    \includegraphics[width=\linewidth]{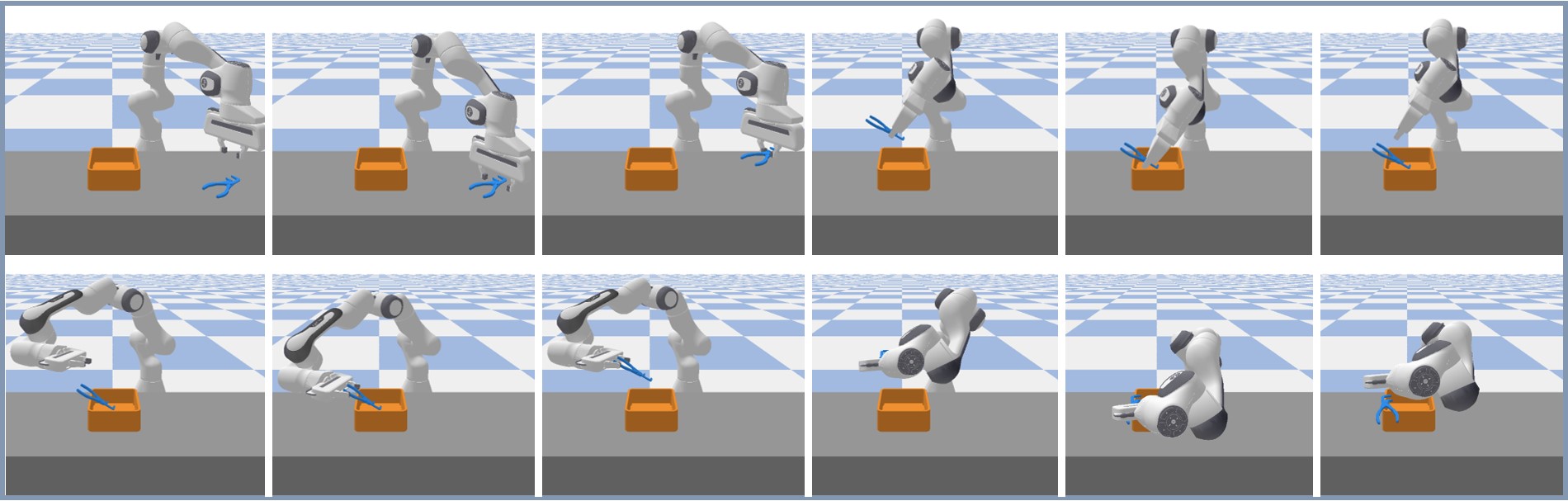} 
    
    \includegraphics[width=\linewidth]{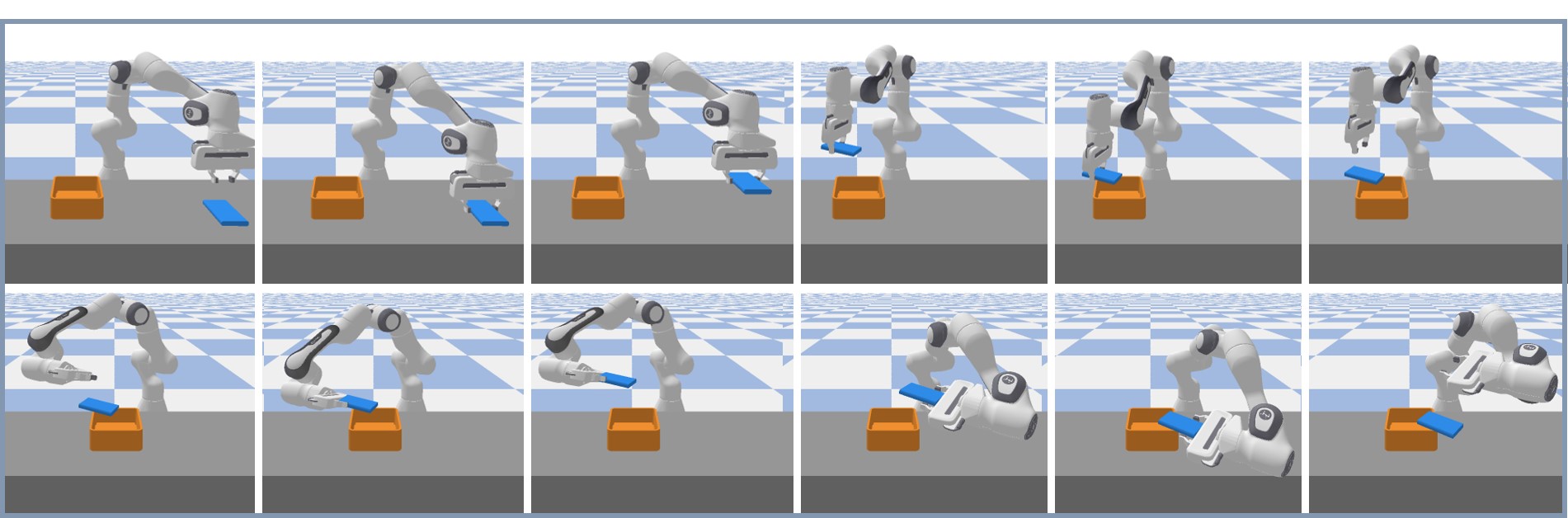} 
    
    \includegraphics[width=\linewidth]{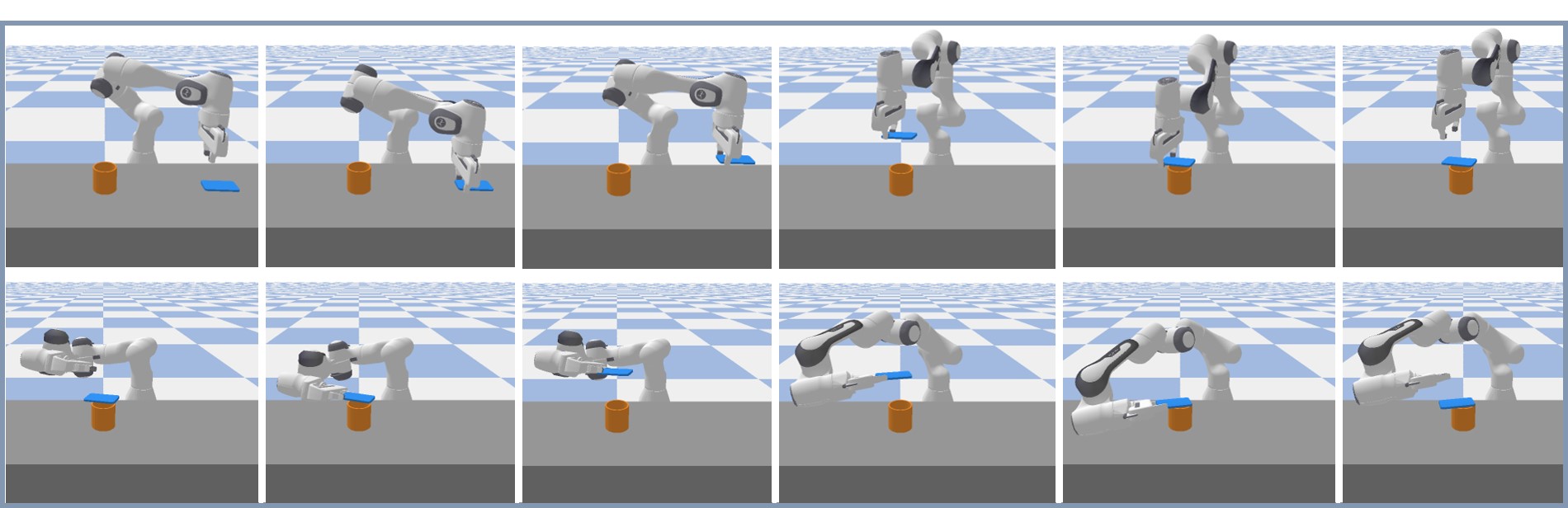} 
    
    \includegraphics[width=\linewidth]{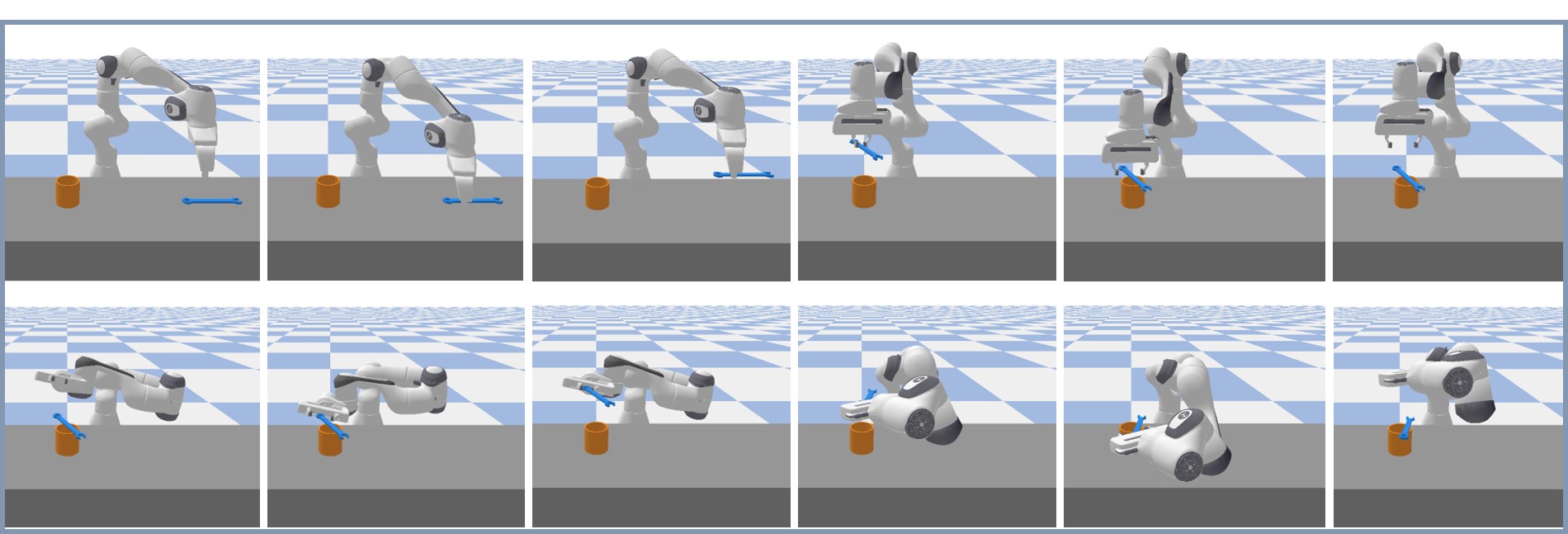}
    
\end{figure*}

\begin{figure*}[ht]
      \centering    
    
\includegraphics[width=\linewidth]{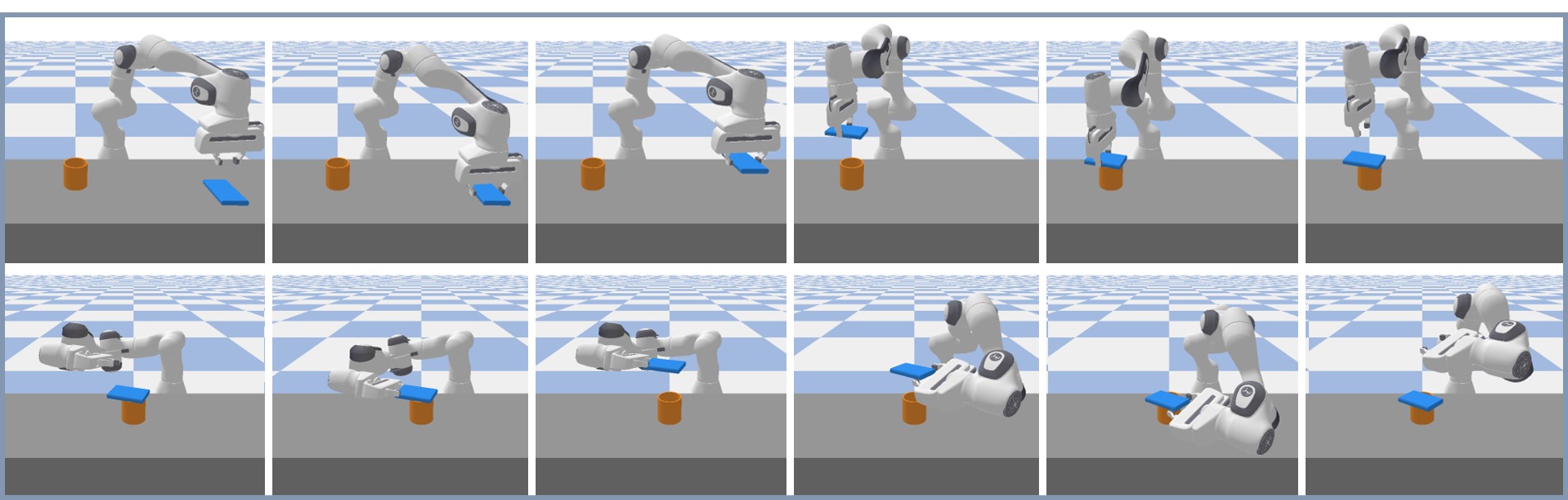}

      \caption{{{Experiments of object flipping. Based on the manipulation graphs, the robot performs sequential pick-and-place operations to flip objects. In each blue box, the images depict two pick-and-place operations performed on an object. The pictures in the first row in each box are the first pick-and-place operation. The pictures in the second row in each box are the second pick-and-place operation. The flipping of each object can be observed from the position change of the gripper. Our approach overcomes the limitation of single pick-and-place in achieving object flipping.}}
      }
      \label{f10}
      
\end{figure*}

\begin{figure*}[ht]
\centering

\includegraphics[width=\linewidth]{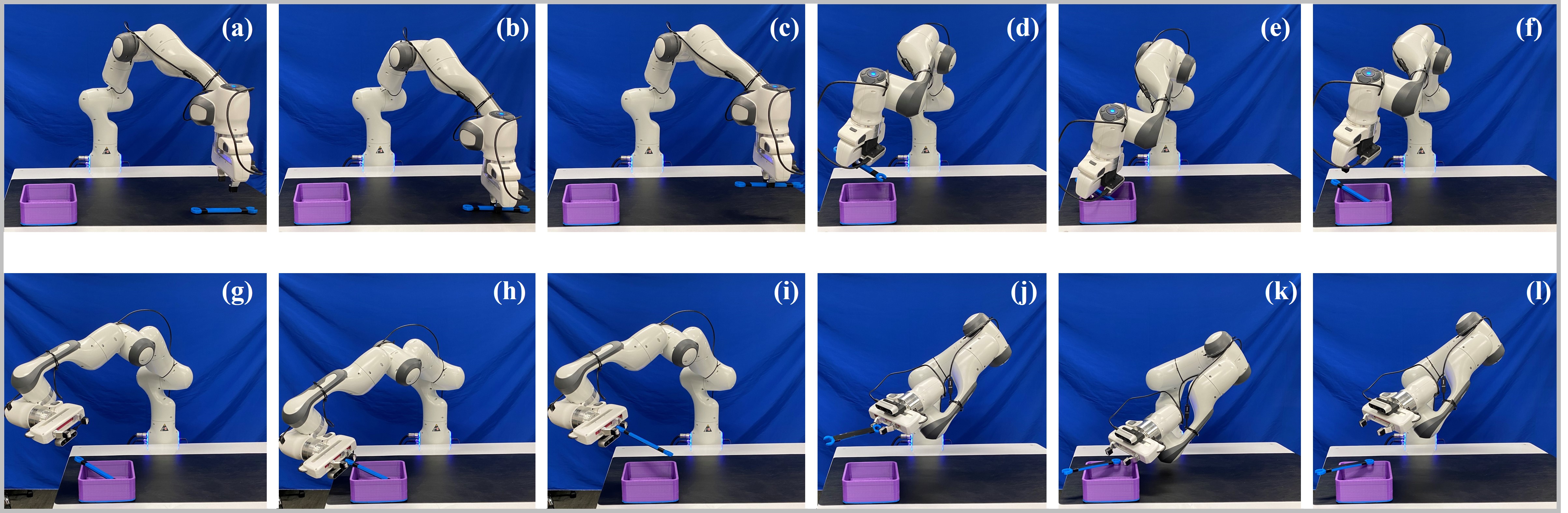}

\caption{A real-world example of flipping objects with stable placements afforded by supports. (a-f) is the first pick-and-place operation. (g-l) is the second pick-and-place operation. The robot performs collision-free sequential pick-and-place operations to flip the object with our designed motion primitives.}

\label{f11_24}
\end{figure*}

\subsection{Evaluations of Predicted Placements}


The comparison results between our method, the state-of-the-art baseline method, and variations of our method on predicting diverse placements of unseen objects are presented in Table \ref{table1}. We compare the prediction results of all methods on 14 pairs of objects with different geometry. In the beginning, each object is positioned randomly beside a support, with two objects separated from each other in the simulation. We leverage point clouds of the object and the supporting environment as inputs for all methods. Subsequently, we place the object at the placements predicted by each method and run each placement simulation process for 5 seconds to allow the object to stabilize under gravity in PyBullet \cite{coumans2020}. The stability of each predicted placement is evaluated by comparing the final position of the object after each simulation process with its predicted placement. To evaluate placements, we set thresholds of 1 cm in three axial directions for the offset and 10 degrees in three axial directions for the angle. We deem a predicted placement to be a stable placement if the translation and orientation changes of the object before and after the simulation fall within all thresholds. As listed in Table \ref{table1}, "rate($\%$)" represents the accuracy rate of the prediction results of a certain method, while "num" represents the number of stable placements predicted by a certain method. 

\subsubsection{Comparison with the State-of-the-art Baseline}

The state-of-the-art baseline method we compare against, L2P \cite{cheng2021learning}, involves using point clouds to predict the placements of objects afforded solely on the support. In contrast, our method predicts placements of objects supported by the support and the table. We emphasize not only the stability but also the diversity of predicted placements to accomplish complex manipulation tasks. The baseline claims to have trained their model using a large amount of simulation data. We use the trained weight provided by them for comparison. When using both methods for inference, we set the number of input noises to 512 and the threshold $S$ for classification to 0.9.

As reported in Table \ref{table1}, our method achieves better accuracy than the baseline on the vast majority of unseen pairs. {{The experiment results show that our method outperforms the state-of-the-art baseline with a 20.2\% improvement in overall accuracy.}} Our method achieves an overall accuracy of 36.9\% (341/924) on the test pairs, whereas the baseline method only achieves 16.7\% (705/4227). We compare the placements predicted by our method with those predicted by the baseline method. The results reveal significant differences in the number of placements for different pairs. The baseline produces no stable placements for almost half of these pairs while generating many placements of the same type for several pairs.
We also present the placements predicted by our method and the baseline in Fig. \ref{f9}. To compare the results, we use images of mesh models captured in simulation to better illustrate the relative positions of objects than using point clouds \cite{cheng2021learning}. 
For pairs of tray102+pliers6, plate1+wrench4, and cup1+wrench4, our method predicts diverse placements, whereas the baseline predictions suffer from inaccuracy caused by penetration. For the pair plate5+pliers6, our method predicts more types of placements than the baseline that generates only one type. More test results of the baseline are available in the link\footnote{https://github.com/pengPeterpeng/Learn-to-regrasp.git}. Our approach shows the generalization ability to novel objects with large variance, producing a wide variety of placements that potentially benefit the robot in complex manipulation tasks.


\subsubsection{Ablation Tests}

We compare our method with its variations. The comparison focuses on two aspects: first, the validity of our loss function design in the pose generation stage (Sec. \ref{secPG}), and second, the role of the pose refinement stage (Sec. \ref{secPR}) in which we learn the dynamics of objects under gravity based on point clouds. 
Regarding the first aspect, we design the first variation method \textit{ours w/ 6d}, which adjusts the loss function to compare 6D poses with the ground truth and keeps the network architecture unchanged. The 6D poses refer to the orientations and corresponding translations of a point cloud transformed to stable placements in the world coordinate frame. The second variation of our method \textit{ours w/o refine} excludes the pose refinement stage, while the requirements of other stages remain unchanged. 

The results in Table \ref{table1} indicate that our method outperforms its variations in terms of both diversity and accuracy of generated placements. Our method generates more stable placements than \textit{ours w/ 6d} on all unseen pairs. The overall accuracy of \textit{ours w/ 6d} is lower than that of our method. This finding is attributed to the fact that orientation and translation are not independent. After the point cloud is rotated with the orientation, its translation changes, and the corresponding translation is added to reach a stable placement in the world coordinate frame. Therefore, it is not easy to directly obtain accurate translation predictions when simultaneously learning the orientations and translations. Furthermore, our method guarantees that the point cloud of the object is around the support in the first stage using the sampling method to obtain translations.
Our method improves overall accuracy from 36.5\% (260/713) of \textit{ours w/o refine} to 36.9\% (341/924). Moreover, an important conclusion is that our method demonstrates high generalization capacity, as it predicts stable placements for some pairs for which \textit{ours w/o refine} does not.

\subsubsection{Limitations}



In this section, we discuss the limitations of our approach. Fig. \ref{failure} illustrates the placements predicted by our approach and their simulation results. The figure presents instances of inaccurately predicted placements, including cases of penetration (e.g., tray102+pliers6 and cup1+wrench4), instances where the difference between the predicted and simulated poses exceeds the threshold (e.g., all pairs in Fig. \ref{failure}), and placements where the object does not make contact with the support (e.g., plate1+wrench4). This underscores the need for further improvement of the classifier model. Simply increasing the score threshold $S$ would result in a decrease in the total number of selected placements. We can enhance the accuracy of classification by improving the performance of the classifier model to encode the refined placements. The classifier can extract continuous features of 3D objects by incorporating implicit representation methods, enabling it to encode features that cannot be obtained through PointNet++ networks.

\subsection{Algorithm Validations}
{{To illustrate the capability of our algorithm in constructing manipulation graphs for object reorientation, we conduct experiments in both simulated and real-world scenarios.}}
In these scenarios, we employ our algorithm to construct manipulation graphs, which facilitate sequential pick-and-place steps for object flipping. 

\subsubsection{Simulation Experiments}
{{
In Fig. \ref{f10}, we present experimental results conducted on a set of previously unseen objects: tray102+pliers6, tray102+disk3, cup1+phone1, cup1+wrench4, and cup1+disk3. As depicted in the figure, the objects to be manipulated are positioned in random initial poses next to their respective support objects. In these scenarios, direct flipping of the objects to attain the desired goal poses with a single pick-and-place step is infeasible. The robot's base is fixed in the environment, and the table obstructs the gripper from grasping objects in a horizontal posture. Consequently, the range of viable robotic motions for object reorientation is restricted.

To address these challenges, we employ our algorithm to construct manipulation graphs. The initial point clouds of each scene are obtained from the fusion of multiple images. Subsequently, our pipeline predicts object placements based on these point clouds. The target poses are given by predicted flip orientations and manually set translations. In the manipulation graphs, predicted placements with elevated scores are prioritized for graph expansion. As a result, we obtain manipulation graphs that incorporate stable placements and sequential pick-and-place steps for object flipping. 
Each experiment in Fig. \ref{f10} shows a sequence of pick-and-place operations performed by the robot to flip the object. The pictures for each row in each box show a pick-and-place operation. Our designed motion primitives, pick-up and place-down, are also shown in these pictures. 
}}

{{Furthermore, we have discerned that the choice of the support exerts influence on object flipping. Given the absence of feasible manipulation sequences in manipulation graphs, object reorientation tasks corresponding to shorter supports (e.g., plate1 and plate5) are not accomplished. Despite diverse stable placements of objects that are afforded by short supports, these placements allow no access to grasp configurations that are conducive to flip objects. This emerges due to the inadequate shapes and sizes of the supports, which hinder placements used by the robot to flip objects via pick-and-place steps.
Consequently, the selection of supports, especially in cluttered scenes, is important to obtain placements for the execution of object reorientation. 
Given language instructions, VoxPoser \cite{huang2023voxposer} is proposed to compose 3D affordance and constraint maps for manipulation tasks, using large language models (LLMs) and vision-language models (VLMs). The values in 3D value maps reflect the entity of interest and entity to avoid. 
In subsequent research, LLMs could be exploited to guide the robot in selecting appropriate supports for object reorientation.
}}

\subsubsection{The Real Robot Experiment}
\label{5.3}

To validate our approach, we present the real-world experiment using the Franka Emika Panda robot equipped with the default parallel-jaw gripper. To enable perception, we mount an Intel RealSense D435 camera on the end-effector and use the eye-in-hand calibration \cite{tsai1989new} to calibrate the transformation matrix between the end-effector and the camera before capturing color and depth images. The communication between the robot and the camera is established through ROS. 
Initially, the object is randomly placed on a table next to a support. Our system constructs the initial scene by fusing a series of color and depth images captured from various angles simultaneously \cite{breyer2020volumetric,Zhou2018}. The point clouds are then sampled from the constructed scene using Farthest Point Sampling \cite{qi2017pointnet++}, and segmentation masks for different instances are determined using \cite{Rusu_ICRA2011_PCL}. Once we obtain the segmented point clouds, we set the world coordinate frame with canonical directions of axes at the center of the support's bottom. We implement our pipeline to predict placements and construct the manipulation graph.

Fig. \ref{f11_24} demonstrates the successful execution of our approach to flip the wrench through two pick-and-place operations in a real-world scenario. 
As per the manipulation graph, the robot first adjusts the grasp direction above the object (a) before reaching it and closing the two-finger gripper to grasp it (b). Next, the robot picks up the object to create enough space to adjust its pose (c) and moves it to a new position (d). Subsequently, the robot places the object in the new placement and opens the gripper (e-f). A similar process is then conducted (g-l) to move the object to the goal pose.
Finally, the robot can reorient the object without collision. The motion planning of the robot is based on the RRT-Connect.
Our approach enables the robot to reorient objects, such as flipping, by utilizing stable placements despite the robot being fixed in the environment. However, we observed changes in the object's pose when the robot grasped it due to the inaccuracy of the contact points predicted by \cite{shao2020unigrasp}. We would further improve the accuracy of our algorithm by substituting this grasp method.

\section{Conclusion}

In this paper, we introduce a novel data-driven pipeline that predicts stable placements of objects supported by extrinsic items. We also propose an algorithm to construct manipulation graphs based on point clouds to accomplish object reorientation tasks. The proposed neural network models in our pipeline are trained with a large-scale dataset that covers various contact cases between objects.
We demonstrate the effectiveness of our approach through experiments in both simulated and real-world environments. 
Our approach outperforms the state-of-the-art baseline method in predicting diverse stable placements. 
In future work, we plan to enhance the performance of the classifier model by incorporating implicit representation methods. In addition, we aim to employ large language models to guide the robot to reorient objects with suitable supports in cluttered environments.

\section*{Acknowledgement}

\thanks{The work is supported by the Shenzhen Key Laboratory of Robotics Perception and Intelligence, Southern University of Science and Technology, Shenzhen 518055, China, under Grant ZDSYS20200810171800001.}

\bibliographystyle{ieeetr}
\bibliography{reference}

\begin{thebibliography}{10}

\bibitem{wan2019regrasp}
W.~Wan, H.~Igawa, K.~Harada, H.~Onda, K.~Nagata, and N.~Yamanobe, ``A regrasp
  planning component for object reorientation,'' {\em Autonomous Robots},
  vol.~43, no.~5, pp.~1101--1115, 2019.

\bibitem{yuan2020design}
S.~Yuan, L.~Shao, C.~L. Yako, A.~Gruebele, and J.~K. Salisbury, ``Design and
  control of roller grasper v2 for in-hand manipulation,'' in {\em 2020
  IEEE/RSJ International Conference on Intelligent Robots and Systems (IROS)},
  pp.~9151--9158, IEEE, 2020.

\bibitem{li2020learning}
T.~Li, K.~Srinivasan, M.~Q.-H. Meng, W.~Yuan, and J.~Bohg, ``Learning
  hierarchical control for robust in-hand manipulation,'' in {\em 2020 IEEE
  International Conference on Robotics and Automation (ICRA)}, pp.~8855--8862,
  IEEE, 2020.

\bibitem{cao2016analyzing}
C.~Cao, W.~Wan, J.~Pan, and K.~Harada, ``Analyzing the utility of a support pin
  in sequential robotic manipulation,'' in {\em 2016 IEEE International
  Conference on Robotics and Automation (ICRA)}, pp.~5499--5504, IEEE, 2016.

\bibitem{ma2018regrasp}
J.~Ma, W.~Wan, K.~Harada, Q.~Zhu, and H.~Liu, ``Regrasp planning using stable
  object poses supported by complex structures,'' {\em IEEE Transactions on
  Cognitive and Developmental Systems}, vol.~11, no.~2, pp.~257--269, 2018.

\bibitem{haustein2019object}
J.~A. Haustein, K.~Hang, J.~Stork, and D.~Kragic, ``Object placement planning
  and optimization for robot manipulators,'' in {\em 2019 IEEE/RSJ
  International Conference on Intelligent Robots and Systems (IROS)},
  pp.~7417--7424, IEEE, 2019.

\bibitem{haustein2019placing}
J.~A. Haustein, S.~Cruciani, R.~Asif, K.~Hang, and D.~Kragic, ``Placing objects
  with prior in-hand manipulation using dexterous manipulation graphs,'' in
  {\em 2019 IEEE-RAS 19th International Conference on Humanoid Robots
  (Humanoids)}, pp.~453--460, IEEE, 2019.

\bibitem{hou2019reorienting}
Y.~Hou, Z.~Jia, and M.~T. Mason, ``Reorienting objects in 3d space using
  pivoting,'' {\em arXiv preprint arXiv:1912.02752}, 2019.

\bibitem{you2021omnihang}
Y.~You, L.~Shao, T.~Migimatsu, and J.~Bohg, ``Omnihang: Learning to hang
  arbitrary objects using contact point correspondences and neural collision
  estimation,'' {\em arXiv preprint arXiv:2103.14283}, 2021.

\bibitem{paxton2021predicting}
C.~Paxton, C.~Xie, T.~Hermans, and D.~Fox, ``Predicting stable configurations
  for semantic placement of novel objects,'' {\em arXiv preprint
  arXiv:2108.12062}, 2021.

\bibitem{sundaralingam2018geometric}
B.~Sundaralingam and T.~Hermans, ``Geometric in-hand regrasp planning:
  Alternating optimization of finger gaits and in-grasp manipulation,'' in {\em
  2018 IEEE International Conference on Robotics and Automation (ICRA)},
  pp.~231--238, IEEE, 2018.

\bibitem{wan2019preparatory}
W.~Wan, K.~Harada, and F.~Kanehiro, ``Preparatory manipulation planning using
  automatically determined single and dual arm,'' {\em IEEE Transactions on
  Industrial Informatics}, vol.~16, no.~1, pp.~442--453, 2019.

\bibitem{cruciani2019dual}
S.~Cruciani, K.~Hang, C.~Smith, and D.~Kragic, ``Dual-arm in-hand manipulation
  and regrasping using dexterous manipulation graphs,'' {\em arXiv preprint
  arXiv:1904.11382}, 2019.

\bibitem{wan2016achieving}
W.~Wan and K.~Harada, ``Achieving high success rate in dual-arm handover using
  large number of candidate grasps, handover heuristics, and hierarchical
  search,'' {\em Advanced Robotics}, vol.~30, no.~17-18, pp.~1111--1125, 2016.

\bibitem{wan2015reorientating}
W.~Wan and K.~Harada, ``Reorientating objects with a gripping hand and a table
  surface,'' in {\em 2015 IEEE-RAS 15th International Conference on Humanoid
  Robots (Humanoids)}, pp.~101--106, IEEE, 2015.

\bibitem{Hou-2018-105030}
Z.~J. Yifan~Hou and M.~T. Mason, ``Fast planning for 3d any-pose-reorienting
  using pivoting,'' in {\em Proceedings of (ICRA) International Conference on
  Robotics and Automation}, pp.~1631 -- 1638, IEEE Robotics and Automation
  Society (RAS), May 2018.

\bibitem{jiang2012learning}
Y.~Jiang, M.~Lim, C.~Zheng, and A.~Saxena, ``Learning to place new objects in a
  scene,'' {\em The International Journal of Robotics Research}, vol.~31,
  no.~9, pp.~1021--1043, 2012.

\bibitem{jiang2012learning2}
Y.~Jiang, C.~Zheng, M.~Lim, and A.~Saxena, ``Learning to place new objects,''
  in {\em 2012 IEEE International Conference on Robotics and Automation},
  pp.~3088--3095, IEEE, 2012.

\bibitem{berscheid2020self}
L.~Berscheid, P.~Mei{\ss}ner, and T.~Kr{\"o}ger, ``Self-supervised learning for
  precise pick-and-place without object model,'' {\em IEEE Robotics and
  Automation Letters}, vol.~5, no.~3, pp.~4828--4835, 2020.

\bibitem{cheng2021learning}
S.~Cheng, K.~Mo, and L.~Shao, ``Learning to regrasp by learning to place,''
  {\em arXiv preprint arXiv:2109.08817}, 2021.

\bibitem{simeonov2020long}
A.~Simeonov, Y.~Du, B.~Kim, F.~R. Hogan, J.~Tenenbaum, P.~Agrawal, and
  A.~Rodriguez, ``A long horizon planning framework for manipulating rigid
  pointcloud objects,'' in {\em Conference on Robot Learning (CoRL)}, 2020.

\bibitem{breyer2020volumetric}
M.~Breyer, J.~J. Chung, L.~Ott, S.~Roland, and N.~Juan, ``Volumetric grasping
  network: Real-time 6 dof grasp detection in clutter,'' in {\em Conference on
  Robot Learning}, 2020.

\bibitem{Rusu_ICRA2011_PCL}
R.~B. Rusu and S.~Cousins, ``{3D is here: Point Cloud Library (PCL)},'' in {\em
  {IEEE International Conference on Robotics and Automation (ICRA)}},
  (Shanghai, China), May 9-13 2011.

\bibitem{veres2017modeling}
M.~Veres, M.~Moussa, and G.~W. Taylor, ``Modeling grasp motor imagery through
  deep conditional generative models,'' {\em IEEE Robotics and Automation
  Letters}, vol.~2, no.~2, pp.~757--764, 2017.

\bibitem{qi2017pointnet++}
C.~R. Qi, L.~Yi, H.~Su, and L.~J. Guibas, ``Pointnet++: Deep hierarchical
  feature learning on point sets in a metric space,'' {\em arXiv preprint
  arXiv:1706.02413}, 2017.

\bibitem{fan2017point}
H.~Fan, H.~Su, and L.~J. Guibas, ``A point set generation network for 3d object
  reconstruction from a single image,'' in {\em Proceedings of the IEEE
  conference on computer vision and pattern recognition}, pp.~605--613, 2017.

\bibitem{shen2022acid}
B.~Shen, Z.~Jiang, C.~Choy, L.~J. Guibas, S.~Savarese, A.~Anandkumar, and
  Y.~Zhu, ``Acid: Action-conditional implicit visual dynamics for deformable
  object manipulation,'' {\em arXiv preprint arXiv:2203.06856}, 2022.

\bibitem{ronneberger2015u}
O.~Ronneberger, P.~Fischer, and T.~Brox, ``U-net: Convolutional networks for
  biomedical image segmentation,'' in {\em Medical Image Computing and
  Computer-Assisted Intervention--MICCAI 2015: 18th International Conference,
  Munich, Germany, October 5-9, 2015, Proceedings, Part III 18}, pp.~234--241,
  Springer, 2015.

\bibitem{peng2020convolutional}
S.~Peng, M.~Niemeyer, L.~Mescheder, M.~Pollefeys, and A.~Geiger,
  ``Convolutional occupancy networks,'' in {\em Computer Vision--ECCV 2020:
  16th European Conference, Glasgow, UK, August 23--28, 2020, Proceedings, Part
  III 16}, pp.~523--540, Springer, 2020.

\bibitem{coumans2020}
E.~Coumans and Y.~Bai, ``Pybullet, a python module for physics simulation for
  games, robotics and machine learning,'' 2016--2020.

\bibitem{shao2020unigrasp}
L.~Shao, F.~Ferreira, M.~Jorda, V.~Nambiar, J.~Luo, E.~Solowjow, J.~A. Ojea,
  O.~Khatib, and J.~Bohg, ``Unigrasp: Learning a unified model to grasp with
  multifingered robotic hands,'' {\em IEEE Robotics and Automation Letters},
  vol.~5, no.~2, pp.~2286--2293, 2020.

\bibitem{huang2023voxposer}
W.~Huang, C.~Wang, R.~Zhang, Y.~Li, J.~Wu, and L.~Fei-Fei, ``Voxposer:
  Composable 3d value maps for robotic manipulation with language models,''
  {\em arXiv preprint arXiv:2307.05973}, 2023.

\bibitem{tsai1989new}
R.~Y. Tsai, R.~K. Lenz, {\em et~al.}, ``A new technique for fully autonomous
  and efficient 3 d robotics hand/eye calibration,'' {\em IEEE Transactions on
  robotics and automation}, vol.~5, no.~3, pp.~345--358, 1989.

\bibitem{Zhou2018}
Q.-Y. Zhou, J.~Park, and V.~Koltun, ``{Open3D}: {A} modern library for {3D}
  data processing,'' {\em arXiv:1801.09847}, 2018.

\end{thebibliography}




\end{document}